\documentclass[10pt,letterpaper]{article}

\usepackage{cogsci}

\cogscifinalcopy

\usepackage{pslatex}
\usepackage{apacite}
\usepackage{natbib}
\usepackage{graphicx}
\usepackage{xcolor}

\title{Psychologically-informed chain-of-thought prompts for metaphor understanding in large language models}

\author{{\large \bf Ben Prystawski$^1$, Paul Thibodeau$^2$, Christopher Potts$^3$, Noah D. Goodman$^{1,4}$} \\
$^1$Department of Psychology, Stanford University \\
$^2$Department of Psychology, Oberlin College\\
$^3$Department of Linguistics, Stanford University \\
$^4$Department of Computer Science, Stanford University \\
{\bf\texttt{\{benpry,pthibod1,cgpotts,ngoodman\}@stanford.edu}}}

\begin{document}

\maketitle

\begin{abstract}
Probabilistic models of language understanding are valuable tools for investigating human language use. However, they need to be hand-designed for a particular domain. In contrast, large language models (LLMs) are trained on text that spans a wide array of domains, but they lack the structure and interpretability of probabilistic models. In this paper, we use chain-of-thought prompts to introduce structures from probabilistic models into LLMs. We explore this approach in the case of metaphor understanding. Our chain-of-thought prompts lead language models to infer latent variables and reason about their relationships in order to choose appropriate paraphrases for metaphors. The latent variables and relationships chosen are informed by theories of metaphor understanding from cognitive psychology. We apply these prompts to the two largest versions of GPT-3 and show that they can improve performance in a paraphrase selection task.

\end{abstract}

\section{Introduction}

How do people understand metaphoric language? For example, when we hear the metaphor \textit{A wish is a rainbow}, how do we know that it is intended figuratively, rather than literally? What are the cognitive representations and processes that enable us to find metaphorical meaning in statements that are literally nonsensical? One approach to answering these questions emerges from a framework that views language understanding as a process of probabilistic inference \citep{chater2006probabilistic, frank2012predicting}. Proponents of this approach have sought to explain language understanding by constructing probabilistic models, which describe latent variables in the mind and the relationships between them. 

Probabilistic models of language understanding have several desirable qualities: they are easily interpretable and can leverage their structure to generalize across different stimuli. However, the latent variables, their possible values, and relationships between them must be hand-designed for a specific task in a specific domain. So far, working with probabilistic models requires us to restrict our domain of study to simple tasks where language use is tightly constrained.

For example, \citet{kao2014nonliteral} developed a probabilistic model of non-literal number word usage which describes human judgements well, but relies on empirical measurements of humans' priors for the specific domains it applies to, like the price of an electric kettle. Many studies of metaphor interpretation span different domains so it is not feasible to design domain-specific priors and models.

In contrast, large language models (LLMs) have implicit knowledge about the wide variety of domains represented in their training corpora. They have been shown capable of inferring and performing tasks based on examples \citep{brown2020language, perez2021true, chan2022data}. Models like GPT-3 have performed well at complex and nuanced tasks like summarizing code \citep{ahmed2022few} and standard benchmarks of general language understanding \citep{wang2019superglue, brown2020language}.

A recent line of work has shown that when language models are prompted to produce intermediate reasoning steps before answering a question, they can perform much better than if they are prompted to give an answer directly \citep[e.g.][]{nye2021show,wei2022chain}. By providing examples of questions and answers that consist of step-by-step reasoning toward a solution, we can induce a model to produce a similar set of steps. This often leads to better performance, especially in tasks that rely on formal reasoning like mathematics and logic puzzles \citep{wei2022chain}. Adding explanations can help language models with general knowledge questions, causal reasoning, and nuanced language understanding tasks \citep{lampinen2022can, zelikman2022star}. \citet{dohan2022language} describe chain-of-thought prompting in language models as inducing \textit{cascades}, which are probabilistic models with string-valued random variables.

In this paper, we connect probabilistic models of language understanding with the large language model GPT-3. We prompt the model to generate rationales that involve identifying the latent variables posited by models of metaphor understanding and describing the relationship between the variables, then choosing the most appropriate paraphrase for the metaphor. This approach enables us to make predictions about metaphor understanding in a wider variety of domains than we could using hand-designed probabilistic models.

\section{Related Work}

A long line of work in cognitive psychology has investigated how people process metaphors and find figurative meaning. One theory emphasizes the primacy of literal interpretations \citep{clark1975understanding}. In this view, people must first derive and reject a literal understanding of a metaphoric sentence before initiating a search for a more appropriate figurative interpretation. That is, a reader encountering the sentence ``A wish is a rainbow'' would first understand it to mean that a wish is a meteorological phenomenon, then they would realize that such an interpretation does not make sense. Finally, they would deliberately construct a more appropriate figurative interpretation---ultimately landing on something like ``A wish is beautiful.'' This account emphasizes the role of deliberate reasoning in interpreting metaphorical language.

Though there is some support for the idea that \textit{novel} metaphors force people to engage in the kind of deliberate cognitive processing suggested by this model, experimental work from cognitive psychology suggests that people usually process metaphorical meaning more automatically \citep{bowdle2005career, holyoak2018metaphor, thibodeau2008productive}. One finding that helps to demonstrate the automaticity of metaphor processing is called the Metaphor Interference Effect \citep{glucksberg1982understanding}. In the task, people are asked to identify statements as literally true or false. Results show that participants quickly identify statements that are literally true as true and \textit{most} statements that are literally false as false. However, they take longer to identify statements that are metaphorically true (but literally false) as false. The additional time required to identify statements that are only metaphorically true as literally false suggests that people automatically generate figurative interpretations of metaphors. Therefore, much of the cognition involved in understanding metaphor is automatic.

Still, response times alone cannot tell us what specifically people infer when they interpret metaphors. Probabilistic models of cognition can help us formalize hypotheses on this question. These models posit that people infer specific latent variables and use them to determine a literal state of the world from a metaphorical statement. For example, \citet{kao2014formalizing} proposed a model in which listeners jointly infer a communicative intent, or \textit{question under discussion} \citep[QUD;][]{roberts2004context} and a literal state of the world. This model accurately predicts the literal properties that humans infer about the referents of metaphors. 

Existing probabilistic models of metaphor understanding have largely been focused on simple domains with clear structures, like inferring literal quantities from non-literal uses of numbers \citep{kao2014nonliteral}. They largely rely on hand-crafted structures and specific priors for their domain. Designing an open-ended probabilistic model that can draw inferences from metaphorical language across domains would be prohibitively difficult as it would involve manually designing background knowledge about many possible domains.

Large language models provide an alternative approach to the problem of specifying background knowledge. Since they are trained on large and diverse corpora of text, they learn rich patterns of association between words. Prior work has shown that LLMs can perform similarly to humans on some classic experiments in cognitive psychology \citep{binz2022using} and exhibit human-like content effects in logical reasoning tasks \citep{dasgupta2022language}. Therefore, it is possible that LLMs are capable of making the human-like associations between words that enable metaphor understanding.

\section{Methods}

Code and data for our analyses are available at \resizebox{\linewidth}{!}{\url{https://github.com/benpry/chain-of-thought-metaphor}}.

\subsection{Data}

We use the Katz corpus \citep{katz1988norms} to evaluate language models' ability to select metaphor paraphrases. This corpus consists of 260 non-literary metaphors and 204 literary metaphors in English. All metaphors are of the form ``[subject] is [object]'' or ``[subjects] are [object(s)]'', where [subject] is being described and [object] is being used metaphorically to attribute properties to the subject. For example, ``a bagpipe is a newborn baby'' and ``clouds are tossed pillows'' are metaphors in the Katz corpus. The dataset contains norms of aspects like comprehensibility, metaphor goodness, and familiarity for each metaphor, collected from human ratings.

In preparing the Katz corpus for our study, we started with all 260 of the non-literary metaphors and the 57 most comprehensible literary metaphors (mean rating $\geq5$ on a $7-$point comprehensibility scale). The literary metaphors were generally more opaque than the non-literary ones (e.g., ``Choppy waves are pale octopi''). 
We filtered out any metaphors that were derogatory or dated (e.g., ``President Nixon is the leper of American politics''). This left us with a total of 280 metaphorical statements (230 non-literary and 50 literary).

We wrote four non-metaphorical paraphrases of each metaphorical statement with different degrees of appropriateness (see Table~\ref{tab:stims}). The best interpretations correctly transferred a relevant  property from the object to the subject (e.g., bagpipes are ``loud'' and so are newborn babies). The second-best paraphrase transferred a less apt property from the object to the subject (e.g., bagpipes are ``delicate'' and so are newborn babies). The third-best paraphrase simply expressed a fact about the subject of the metaphor which was irrelevant to the object (e.g., ``A bagpipe is a musical instrument.''). Finally, the worst paraphrases were constructed to express the opposite interpretation of the best paraphrases (e.g., that bagpipes are ``quiet'' and so are newborn babies). Each of these paraphrases was associated with an appropriateness score that ranged from 1 (worst) to 4 (best).

\begin{table}
    \centering
    \begin{tabular}{p{1.5cm}p{5.6cm}} \hline
         Metaphor &  A bagpipe is a newborn baby.\\ \hline
         Best (4) & A bagpipe is loud.\\
         3 & A bagpipe is delicate.\\
         2 & A bagpipe is a musical instrument.\\
         Worst (1) & A bagpipe is quiet. \\ \hline
    \end{tabular}
    \caption{Example metaphoric statement and candidate paraphrases for the model to choose from.}
    \label{tab:stims}
\end{table}


We randomly divided the corpus into training, development, and test sets. The training set consists of 30 metaphors and was used to produce examples of rationales and correct answers to the paraphrasing task for the prompts. The development set consists of 100 metaphors and was used when tuning and refining prompts. The test set consists of the remaining 150 metaphors and was used in the final evaluation (25 literary and 125 non-literary metaphors). 

\subsection{Models}

We used OpenAI's GPT-3 for our analyses \citep{brown2020language}. In particular, we used two different models in the GPT-3 family: DaVinci fine-tuned to follow instructions (text-davinci-002) and Curie (text-curie-001). DaVinci is the largest model, while Curie is one step smaller in size. While OpenAI does not disclose the precise parameter counts of these models, DaVinci performs similarly on common benchmarks to the 175 billion parameter version of GPT-3 and Curie performs similarly to the 6.7 billion parameter version reported by \citet{brown2020language} \citep{gao2021sizes}. All models were queried using the OpenAI API for Python. We set the temperature parameter to $.2$ for all of our analyses to reduce randomness in our results.

\subsection{Prompt Types}

Our prompts each consisted of an instructional sentence followed by 10 examples. The instructional sentence was ``Choose the most appropriate paraphrase of the first sentence.'' After this, the examples contained a metaphoric sentence in quotation marks followed by the four paraphrase options. The paraphrases had letters a through d before them. After the paraphrases, there was a paragraph containing the answer and a possible rationale. After the 10 examples, the prompt contained another metaphorical sentence and four options for paraphrases. Metaphors used for examples were selected from the training set as those that the authors thought could most clearly be explained using rationales.

We tested two types of psychologically-informed rationales: QUD and similarity (see Table~\ref{tab:prompts}). Each was inspired by current theorizing about human metaphor processing from the cognitive psychological literature. Prompts were tuned to maximize their performance on the development set.

\begin{table}
    \centering
    \begin{tabular}{p{1.5cm}|p{5.6cm}} \hline
         QUD &  ``The speaker is addressing the question ``How does a bagpipe sound?''
The speaker answers this question by comparing a bagpipe to a newborn baby.
A newborn baby is loud, so the speaker is saying a) a bagpipe is loud.''\\ \hline
         Similarity & ``A newborn baby is loud.
A bagpipe is also loud, so the speaker is saying a) a bagpipe is loud.''\\ \hline
    \end{tabular}
    \caption{Examples of the different prompt types for the metaphoric statement ``A bagpipe is a newborn baby.''}
    \label{tab:prompts}
\end{table}

QUD prompts were based on the concept of questions under discussion \citep{roberts2004context}. In this account, when a speaker uses a non-conventionalized metaphor, they are implicitly addressing a question about the salient feature being applied to the subject. For example, when a speaker says ``a bagpipe is a newborn baby'' they might be addressing the question ``how loud is a bagpipe?'' Questions under discussion have previously been incorporated into probabilistic models of language understanding, so they are a suitable choice for this task \citep[e.g.][]{qing2016rational}. In QUD prompts, the first part of the rationale identifies the question the speaker is addressing. The second part identifies the subject and the object. The third part highlights the property of the object that is being applied to the subject. Finally, the prompt selects the most appropriate paraphrase.

Similarity prompts were based on comparison accounts of metaphor \citep[e.g.][]{bowdle2005career}, which emphasize finding properties in common between the subject and object of a metaphor. These prompts consist of three parts. First, a sentence highlights the relevant property of the object. Next, a sentence describes the subject's relation to that property (e.g. ``bagpipes are also loud''). Finally, the prompt selects the most appropriate paraphrase.

\begin{figure}
    \centering
    \includegraphics[width=\linewidth]{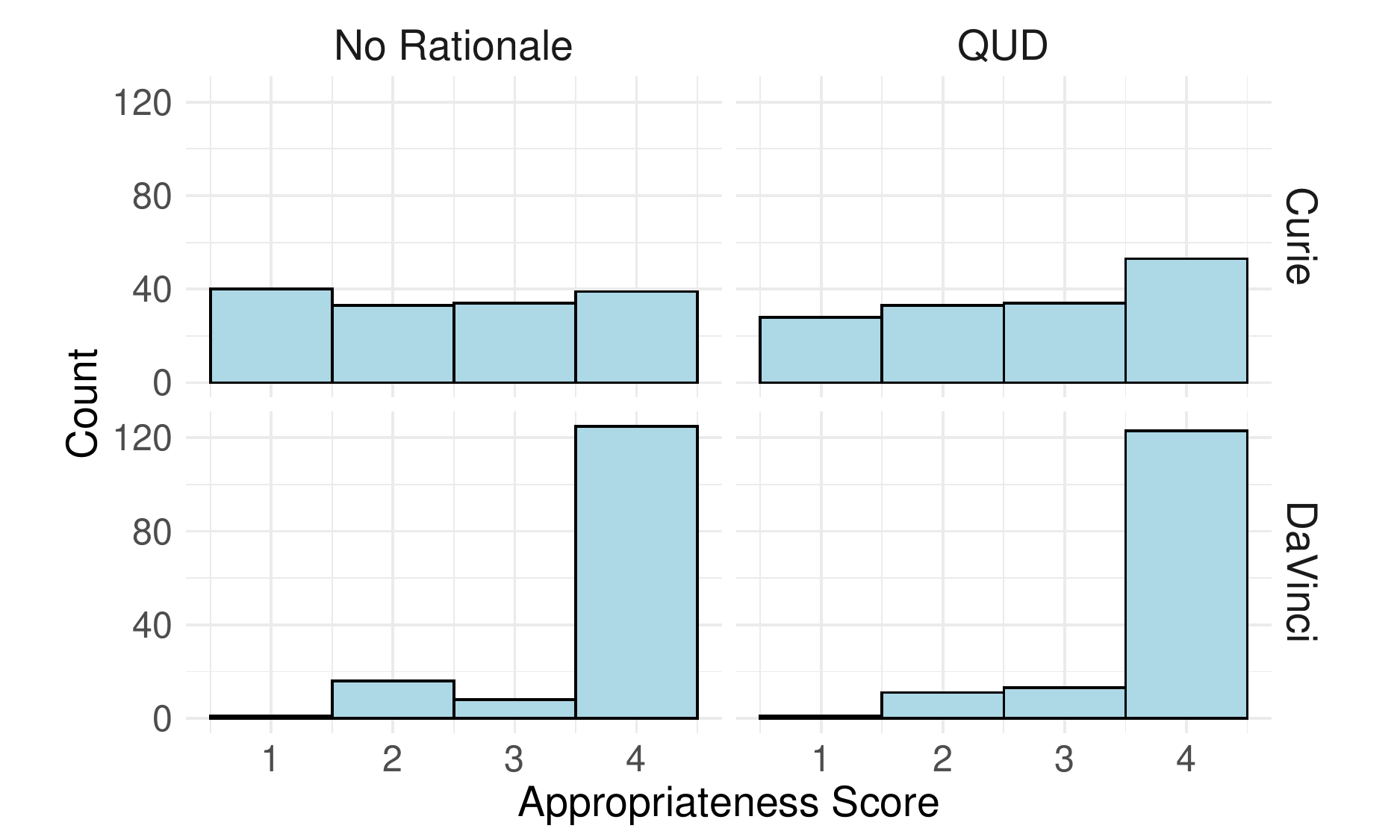}
    \caption{Histograms of paragraph appropriateness distributions for GPT-3 Curie (top) and DaVinci (bottom) with both no rationales (left) and QUD rationales (right).}
    \label{fig:appropriateness_distributions}
\end{figure}

\subsection{Metrics and Evaluation}

\begin{table*}
    \centering
    \resizebox{\linewidth}{!}{\begin{tabular}{cllllll}
        \hline
        Model & Options Only & No Rationale & Non-Explanation & Subject-Object & QUD & Similarity\\\hline
        Curie & 2.52 [2.33, 2.71] & 2.49 [2.31, 2.68] & 2.55 [2.34, 2.74] & 2.63 [2.42, 2.84] & 2.76 [2.57, 2.94] & 2.72 [2.54, 2.91] \\
        DaVinci & 2.73 [2.57, 2.89] & 3.71 [3.60, 3.81] & 3.58 [3.45, 3.71] & 3.84 [3.76, 3.92] & 3.74 [3.64, 3.84] & 3.68 [3.55, 3.79] \\
        \hline
    \end{tabular}}
    \caption{Mean appropriateness scores and bootstrapped 95\% confidence intervals for the paraphrases chosen by each model (rows) using each type of prompt (columns) on the test metaphors.}
    \label{tab:paraphrase_performances}
\end{table*}

\begin{table*}
    \centering
    \hspace*{-2em}\begin{tabular}{cllllll}
        \hline
        Model & Options Only & No Rationale & Non-Explanation & Subject-Object & QUD & Similarity\\\hline
        Curie & 9 & 4 & 22 & 47 & 2 & 12 \\
        DaVinci & 0 & 0 & 1 & 2 & 2 & 5 \\
        \hline
    \end{tabular}
    \caption{Number of responses which failed to be parsed to a multiple-choice answer for each model (rows) and prompt type (columns) on the 150 test metaphors.}
    \label{tab:non_parsed}
\end{table*}

We extracted multiple-choice answers from the model's responses using a regular expression. The examples in each prompt either contained the phrase ``the answer is [a-d])'' or ``the speaker is saying [a-d])''. We took the letter in this phrase to be the model's answer. If we could not match either of these phrases in the model response, we considered the response to be invalid and excluded it from the analysis.

The main metric we use in evaluating model responses is the appropriateness score of the paraphrases chosen by the model. This score ranges between 1 (least appropriate) to 4 (most appropriate). We use this metric as a response variable in Bayesian cumulative logistic regression, as implemented in the BRMS package for R \citep{burkner2017brms}. We also use Pearson correlations to compare across prompting conditions.

To better understand why the model performs how it does, we use ratings of psycholinguistic features of the metaphors that were published with the corpus \citep{katz1988norms}. We are particularly interested in whether performance will differ as a function of the \textit{familiarity} of the metaphors. If a model-prompt combination performs better on more familiar metaphors, it probably relies on having seen that metaphor (or similar metaphors) in training. A combination performing similarly well on familiar and unfamiliar metaphors indicates that it reasons systematically about the metaphor.

\subsection{Baselines}

We compare the performance of our rationale prompts against five baselines: random chance, No Rationales, True Non-explanations, Subject-Object rationales, and Options Only.

The first baseline is random chance. If we simply selected paraphrases at random, we would achieve a mean appropriateness score of 2.5. We compare performance against the random baseline by randomly sampling appropriateness scores from $\{1, 2, 3, 4\}$ an equal number of times to the number of model responses. We did this 10,000 times to compute a distribution over means. The $p$-value is the proportion of means that are at least as different from the random mean as the score of the model we are comparing against chance.

The second baseline is 10-shot prompting with no rationales, meaning we provided 10 examples of metaphors and answers with the most appropriate paraphrase. This lets us test how well a model performs with examples of the task being performed correctly, but no intermediate reasoning steps. 

Third, we created a baseline of true non-explanations, similar to a baseline used in prior work \citep{lampinen2022can}. These prompts were true and domain-relevant statements, but they did not explain the meaning of the metaphor. They first stated a relevant fact (e.g., that bagpipes originated in Scotland), then selected the most appropriate paraphrase.

We also created a baseline that identifies the subject and the object of the metaphor, but does not reason about the property transferred from the object to the subject. This baseline enables us to measure how much of the improvement resulting from rationales is simply because they decompose the metaphorical sentence into a subject and object.

Finally, we created an options only baseline consisting of the possible paraphrases without the metaphorical sentence they are paraphrasing. This lets us verify that the models actually need to see the metaphor to do well on the task rather than relying on superficial features of the paraphrases.

All baseline prompts used the same 10 example metaphors as the rationale prompts, presented in a random order.
 
\section{Results} 

The first notable result is the stark difference between the two GPT-3 models. While Curie performs no better than chance without rationales, DaVinci performs strongly. Without any rationale prompting, GPT-3 DaVinci achieves a mean appropriateness of 3.71 out of 4 on the metaphor paraphrase task while GPT-3 Curie is at chance. Table~\ref{tab:paraphrase_performances} shows the mean appropriateness score for each model with each rationale type on the test set of the Katz corpus. In the following subsections we compare model performance across prompt conditions for the Curie and DaVinci models.

\subsection{Curie}


Some of the rationales led Curie to generate more invalid responses. Only 4 of the 150 responses to test prompts with No Rationales were invalid, but this number increased to 47 with Subject-Object prompts and decreased to 2 with QUD prompts. Table~\ref{tab:non_parsed} shows the number of responses which did not contain a letter between a and d followed by a right parenthesis and thus could not be parsed. 

With No Rationales, Curie's performance did not differ significantly from chance ($p = .97$). The difference was also not significant in the True Non-Explanation ($p = .61$) Subject-Object ($p = .21$), and Options Only ($p = .88$) baselines. QUD prompts led the model to perform significantly better than chance ($p = .005$), as did Similarity prompts ($p = .02$). The difference between performance with QUD prompts and No Rationales was close to significant in a Bayesian ordinal regression, but still not statistically significant ($\beta = .41, [-.02  , .84]$). The top row of Figure~\ref{fig:appropriateness_distributions} shows the appropriateness distributions for GPT-3 Curie both with No Rationales and QUD prompts. The QUD prompts shift the appropriateness distribution away from 1 and toward 4.

We ran a series of correlational analyses to test how similarly Curie responded across the six prompting conditions. As shown in Table~\ref{tab:curie_correl}, we found that the model's performance on the baseline prompts was significantly correlated with its performance given the Similarity rationales, but only performance with True Non-Explanations was significantly correlated with QUD rationale performance. This suggests that the QUD-based prompts influenced the model to perform qualitatively differently than the other types of prompts.

\subsection{DaVinci}

\begin{table}
    \centering
    \begin{tabular}{lccccc}
        \hline
         & 2. & 3. & 4. & 5. & 6. \\ \hline
         1. Options Only & .35** & .27* & .09 & -.07 & .12 \\
         2. No rationale & & .52** & .23* & .08 & .36** \\
         3. Non-explain & & & .29* & .31** & .48** \\
         4. Subj-Obj. & & & & .07 & .32 ** \\
         5. QUD & & & & & .30 ** \\
         6. Similarity & & & & & \\
         \hline
    \end{tabular}
    \caption{Pearson correlations between response appropriateness scores given by Curie across six prompting conditions. Asterisks indicate statistical significance at the *$p<.05$ and **$p<.001$ levels.}
    \label{tab:curie_correl}
\end{table}


DaVinci was very good at choosing metaphor paraphrases without any rationales, achieving a mean appropriateness score of $3.71$ in the No Rationale baseline condition. However, including the metaphor was necessary to perform well at choosing paraphrases, as the mean appropriateness score was only 2.72 in the Options Only baseline. All prompts led this model to perform significantly above chance ($p=.013$ for Options Only baseline, $p < .001$ for all other prompt types). Mean appropriateness was highest with Subject-Object prompts ($3.84$). 

The bottom row of Figure~\ref{fig:appropriateness_distributions} shows the full appropriateness distributions for GPT-3 DaVinci with no rationale prompts and QUD rationale prompts. Both distributions are heavily weighted toward 4, but there are slightly more responses with appropriateness 3 than 2 when using QUD prompts.

As with Curie, we ran a series of correlational analyses to test how similarly the model responded across the six prompting conditions. As shown in Table~\ref{tab:davinci_correl}, the model's behavior across all prompting conditions was correlated, which is not surprising since the model performed well with all types of prompts. Nevertheless, we found informative variability in the strength of the correlations between prompting conditions. Correlations were highest between the three baselines, indicating that they largely succeeded and failed at the same metaphors. Correlations between the QUD and Similarity prompts were weaker both with the baselines and with each other. This suggests that the psychologically-informed prompts represent approaches to metaphor understanding that are different from both the baselines and each other. 

\begin{table}
    \centering
    \begin{tabular}{lccccc}
        \hline
         & 2. & 3. & 4. & 5. & 6. \\ \hline
         1. Options Only & .12 & .12 & .06 & .15 & .08 \\
         2. No rationale && .59** & .53** & .42** & .25* \\
         3. Non-explain &&& .51** & .27** & .47**  \\
         4. Subj-Obj. &&&& .27** & .30** \\
         5. QUD &&&&& .22*  \\
         6. Similarity &&&&&\\
         \hline
    \end{tabular}
    \caption{Pearson correlations between responses given by DaVinci across six prompting conditions. Asterisks indicate statistical significance at the *$p<.05$ and **$p<.001$ levels.}
    \label{tab:davinci_correl}
\end{table}

\subsection{Familiarity and Reasoning Success}

\begin{table*}
    \centering
    \resizebox{\linewidth}{!}{\begin{tabular}{cllllll}
    \hline
    Model & Options Only & No Rationale & True Non-Explanation & Subject-Object & QUD & Similarity  \\\hline
    Curie & 0.09 [-0.14, 0.32] & 0.13 [-0.11, 0.35] & 0.16 [-0.07, 0.39] & -0.08 [-0.35, 0.2] & 0.04 [-0.19, 0.27] & -0.04 [-0.28, 0.2]  \\
    DaVinci & 0.16 [-0.08, 0.41] & 0.35 [0.03, 0.73] & 0.19 [-0.09, 0.49] & 0.2 [-0.2, 0.61] & 0.2 [-0.14, 0.56] & 0.08 [-0.24, 0.37] \\ \hline
    \end{tabular}}
    \caption{Point estimates and 95\% credible intervals for the effect of familiarity on metaphor appropriateness in Bayesian ordinal regression for each combination of model and prompt type.}
    \label{tab:familiarity_effects}
\end{table*}

\begin{table*}
    \centering
    \begin{tabular}{cllllll}
    \hline
    Model & Options Only & No Rationale & True Non-Explanation & Subject-Object & QUD & Similarity  \\ \hline
    Curie & 0.06& 0.24& 0.2& -0.15& 0.07& -0.11 \\
    DaVinci & 0.33& 0.13& 0.2& 0.03& 0.09& -0.01 \\
    \hline
    \end{tabular}
    \caption{Differences between mean appropriateness of chosen paraphrases for the 30 most and least familiar metaphors for each combination of model and prompt type.}
    \label{tab:top_vs_bottom_30}
\end{table*}

We used the psycholinguistic norms in the Katz corpus to analyze how the models' behavior in different prompting conditions relates to aspects of metaphor, focusing particularly on the role of \textit{familiarity}. Metaphors that humans find more familiar likely occur more often in natural language, or are similar to metaphors that occur often, so large language models are more likely to have seen them in training. Understanding the relationship between familiarity and paraphrasing success can help us learn whether language models rely on seeing particular metaphors in training or generalize successfully to understand novel metaphors.

With No Rationales, we found a significant effect of familiarity on paraphrase appropriateness in a Bayesian ordinal regression for GPT-3 DaVinci ($\beta = .36, [.01, .73]$) but not Curie ($\beta = .13, [-.10, .36]$). This effect was not significant for either model when rationales were used. Effects for all combinations of model and prompt type are reported in Table~\ref{tab:familiarity_effects}. This suggests that GPT-3 DaVinci might succeed without rationales by relying on similar metaphors in its training corpus. Adding rationales could make DaVinci less reliant on familiarity. The rationales could encourage language models to break down metaphorical statements into parts and reason systematically about their relationships. This mirrors the finding in cognitive psychology that people tend to understand familiar metaphors automatically, but often need to reason deliberately to understand novel metaphors.

Still, GPT-3 DaVinci performs very well on the paraphrasing task without rationales, even on the more novel metaphors. It would be useful to have a corpus of especially difficult and unfamiliar metaphors to test DaVinci further.


Table~\ref{tab:top_vs_bottom_30} shows the differences in mean appropriateness between the 30 most and least familiar metaphors. This gives an intuitive picture of the magnitude of the difference by familiarity. On performance without rationales we see a small effect of familiarity, bigger for Curie. With QUD rationales there is a much smaller effect, suggesting that reasoning helps close the gap by making more novel metaphors interpretable.
However, this metric is still somewhat noisy as some model-prompt combinations that show an overall increase in appropriateness with familiarity show the opposite effect when focusing on just the 30 most and least familiar metaphors.

\subsection{Error Analysis}

We found that the models generated two main kinds of errors. The following examples come from the condition that used QUD-based prompts with GPT-3 Curie, but they reflect the kinds of mistakes that the model made in the other prompt conditions as well. One was that it seemed to lack an understanding of semantic nuance in some cases. For example, one of the metaphors in the corpus was ``Memories are the roots that clutch.'' The model initially made reasonable inferences in response to this metaphor: ``The speaker is addressing the question ``How do memories affect the mind?'' The speaker answers this question by comparing memories to roots. Roots hold the soil in place'' However, it concluded, ``...so the speaker is saying b) Memories are long.'' In fact, the best paraphrase for that metaphor read ``Memories are stabilizing.'' The model missed the connection between holding soil in place and stabilizing a mind.

The second kind of mistake that the model made seemed more random. For example, one metaphor in the corpus was ``Love is a flower.'' The model responded: ``The speaker is addressing the question ``How does love feel?'' The speaker answers this question by comparing love to a flower. A flower is beautiful, so the speaker is saying c) Love is an emotion.'' The most appropriate paraphrase was ``Love is beautiful and grows.'' The model produced the word ``beautiful,'' but failed to link it to the paraphrase option containing ``beautiful.'' It is not clear why this happens. 

It is important to note that even when the model reached the wrong conclusion, it would almost always demonstrate some appropriate reasoning.

\section{Discussion} 

The primary goal in this paper was to test whether introducing rationales that reason about the latent variables posited by probabilistic models of language understanding could improve the capacity of large language models to choose appropriate paraphrases for metaphors. We developed two kinds of theoretically-motivated prompts based on psychological models of metaphor comprehension and evaluated them on a paraphrasing task.

The first takeaway from our investigation is that some LLMs are already good at the task of matching metaphors to appropriate paraphrases. The paraphrases that DaVinci chose were usually the most appropriate (mean = 3.71/4.0 with no rationale). However, it is also important to emphasize the variability in LLMs' ability to succeed in this task. Curie performed close to chance (mean = 2.49/4.0) with no rationale. 

The second main takeaway is that language models can improve at paraphrasing metaphors if they are prompted to generate rationales that reason about relationships between latent variables. We found that Curie benefited the most from prompting that highlighted a question under discussion, and that DaVinci benefited the most from prompting that simply identified the subject and object of the metaphor. DaVinci's strong performance in the Subject-Object baseline suggests that it might have some difficulty understanding the syntax of a metaphor without rationales that lead it to do so explicitly.

Finally, adding rationales to prompts might reduce GPT-3 DaVinci's reliance on familiarity. There was a significant effect of metaphor familiarity on paraphrase appropriateness for DaVinci with No Rationales, but not when rationales were used. There may be a connection between this finding in DaVinci and the psychological finding that people only engage in deliberate reasoning when encountering novel (i.e., unfamiliar) metaphors. If chain-of-thought prompts lead language models through a process analogous to human reasoning, then this process may be especially helpful when understanding unfamiliar metaphors. Still, the fact that DaVinci performed very well on most metaphors in our corpus and there was not a significant effect of familiarity in other baseline prompting conditions suggests that we should be cautious in interpreting the effect of familiarity.

A limitation of this analysis is that the authors wrote the paraphrases for the metaphors in the Katz corpus. The paraphrases and prompts reflect the authors' understanding of the metaphors rather than a representative sample of English speakers. Furthermore, the familiarity ratings in the Katz corpus reflect participants' intuitions about metaphors' familiarity, which might differ from the actual familiarity of the metaphors. Previous work has questioned the validity of subjective ratings of metaphors \citep{thibodeau2017subjective}.

Future work should test the hypothesis that rationales make GPT-3 DaVinci less reliant on familiarity in metaphor understanding. It would be informative to develop a dataset of very difficult and novel metaphors. We could evaluate DaVinci's performance on these metaphors. If it does poorly on the very novel metaphors, we could test whether rationales can help its performance. This would enable us both to understand the limits of GPT-3 DaVinci better and understand the effect of rationales on novel metaphor comprehension.

\bibliographystyle{apacite}

\setlength{\bibleftmargin}{.125in}
\setlength{\bibindent}{-\bibleftmargin}

\bibliography{citations}

\end{document}